%% file: main.tex
\documentclass[9pt, a4paper, logo, twocolumn, copyright, nonumbering]{googledeepmind}

\usepackage[authoryear, sort&compress, round]{natbib}
\usepackage{subcaption}
\usepackage{caption}
\title{RecurrentGemma: Moving Past Transformers for Efficient Open Language Models}

\correspondingauthor{[botev, sohamde, slsmith, anushanf]@google.com}

\reportnumber{001} 
\usepackage{stfloats}

\author[1]{{Griffin}}
\author[1]{{RLHF}}
\author[1]{{Gemma Teams}}

\affil[1]{Google DeepMind. Please see contributors and acknowledgements section for full author list.}

\begin{abstract}
We introduce RecurrentGemma, a family of open language models which uses Google's novel Griffin architecture. Griffin combines linear recurrences with local attention to achieve excellent performance on language. It has a fixed-sized state, which reduces memory use and enables efficient inference on long sequences. We provide two sizes of models, containing 2B and 9B parameters, and provide pre-trained and instruction tuned variants for both. Our models achieve comparable performance to similarly-sized Gemma baselines despite being trained on fewer tokens.
\end{abstract}

\begin{document}

\maketitle

\section{Introduction}

We present RecurrentGemma, a family of open models based on the Griffin architecture \citep{de2024griffin}. This architecture eschews global attention, instead modelling the sequence through a mixture of linear recurrences \citep{gu2021efficiently, orvieto2023resurrecting} and local attention \citep{beltagy2020longformer}. 
We provide two sizes of RecurrentGemma, with 2B and 9B parameters, both trained on 2T tokens. Our models achieve superb performance on a range of downstream tasks, competitive with the Gemma models \citep{gemma2024}, an open transformer model family based on insights from Gemini \citep{geminiteam2023gemini}.

To perform inference, transformers must retrieve the KV cache and load it into device memory. This KV cache grows linearly with sequence length. Although one can reduce the cache size by using local attention \citep{beltagy2020longformer}, this comes at the cost of reduced performance. In contrast, RecurrentGemma compresses input sequences into a fixed-size state without sacrificing performance. This reduces memory use and enables efficient inference on long sequences. We verify below that RecurrentGemma models achieve faster inference than Gemma models.

For each model size, we are releasing both a pre-trained checkpoint and an instruction tuned checkpoint fine-tuned for instruction-following and dialogue.\footnote{\url{https://github.com/google-deepmind/recurrentgemma}} We are also releasing efficient JAX code to evaluate and fine-tune our models \citep{bradbury2018jax}, including a specialized Pallas kernel to perform the linear recurrence on TPUs. We provide a reference PyTorch implementation as well.

\section{Model architecture}
We make only a single modification to the Griffin architecture \citep{de2024griffin}, which is to multiply the input embeddings by a constant equal to the square root of model width. The input and output embeddings are tied, but this factor is not applied to the output. A similar multiplicative factor appears in Gemma \citep{gemma2024}. 
We define the key model hyper-parameters for both RecurrentGemma-2B and RecurrentGemma-9B in Table \ref{table:hypers}, and defer the reader to \citet{de2024griffin} for exact details on the overall architecture. 

Note that we do not apply weight decay to the parameters of the recurrent (RG-LRU) layers during training. Additionally when backpropagating through the square root operation in the recurrent layers, we always clip the derivative to a maximum value of 1000 for stability.

\begin{table}[t!]
\caption{Key model hyper-parameters. See Griffin paper \citep{de2024griffin} for model definition.}
\begin{tabular}{lll}
\toprule
\textbf{RecurrentGemma-} & \textbf{2B} & \textbf{9B} \\
\midrule
\textbf{Total params} & 2.68B & 8.58B \\
\textbf{Non-Embedding params} & 2.03B & 7.53B \\
\textbf{Embedding params} & 0.65B & 1.05B \\ \midrule
\textbf{Vocabulary size} & 256k & 256k \\
\textbf{Model width} & 2560 & 4096 \\
\textbf{RNN width} & 2560 & 4096 \\
\textbf{MLP expansion factor} & 3 & 3 \\
\textbf{Depth} & 26 & 38 \\
\textbf{Attention heads} & 10 & 16 \\
\textbf{Local attention window size} & 2048 & 2048 \\
\bottomrule
\end{tabular} \label{table:hypers}
\end{table}

\section{Training details}

\begin{table*}[t]
\vspace{-2mm}
\centering
\caption{Academic benchmark results, compared to the Gemma models. Note that Gemma-7B contains a similar total number of parameters to RecurrentGemma-9B (after accounting for embedding layers). Gemma-2B was trained on 3T tokens and Gemma-7B was trained on 6T tokens, while both RecurrentGemma-2B and RecurrentGemma-9B were trained on 2T tokens.}
\begin{tabular}{l c  c c  c c}
\toprule
& & \multicolumn{2}{c}{Gemma} & \multicolumn{2}{c}{RecurrentGemma} \\
\cmidrule(l{3pt}r{3pt}){3-4}\cmidrule(l{3pt}r{3pt}){5-6}
Benchmark & Metric & 2B & 7B & 2B & 9B \\ \midrule
MMLU & 5-shot, top-1 & 42.3 & 64.3 & 38.4 & 60.5 \\ \hline
HellaSwag & 0-shot & 71.4 & 81.2 & 71.0 & 80.4 \\
PIQA & 0-shot & 77.3 & 81.2 & 78.5 & 81.3 \\
SIQA & 0-shot & 49.7 & 51.8 & 51.8 & 52.3 \\
Boolq & 0-shot & 69.4 & 83.2 & 71.3 & 80.3 \\
Winogrande & partial scoring & 65.4 & 72.3 & 67.8 & 73.6 \\
CQA & 7-shot & 65.3 & 71.3 & 63.7 & 73.2 \\
OBQA &  & 47.8 & 52.8 & 47.2 & 51.8 \\
ARC-e &  & 73.2 & 81.5 & 72.9 & 78.8 \\
ARC-c &  & 42.1 & 53.2 & 42.3 & 52.0 \\ \hline
TriviaQA & 5-shot & 53.2 &  63.4 & 52.5 & 70.5 \\
NQ & 5-shot & 12.5 &  23.0 & 11.5 & 21.7 \\ \hline
HumanEval & pass@1 & 22.0 & 32.3 & 21.3 & 31.1 \\
MBPP & 3-shot & 29.2 & 44.4 & 28.8 & 42.0 \\
GSM8K & maj@1 & 17.7 & 46.4 & 13.4 & 42.6 \\
MATH & 4-shot & 11.8 & 24.3 & 11.0 & 23.8 \\ \hline
AGIEval &  & 24.2 & 41.7 & 23.8 & 39.3 \\
BBH &  & 35.2 & 55.1 & 35.3 & 55.2 \\ \midrule
Average &  & 45.0 & 56.9 & 44.6 & 56.1 \\ \bottomrule
 & \multicolumn{1}{l}{} & \multicolumn{1}{l}{} & \multicolumn{1}{l}{}
\end{tabular} \label{table:evals}
\end{table*}

\subsection{Pre-training}

We train on sequences of 8192 tokens. We use the same pre-training data as the Gemma models, which comprises primarily English data from web documents, mathematics and code. This dataset was filtered to reduce the risk of unwanted or unsafe utterances, and to filter out personal or sensitive data as well as to filter out all evaluation sets from our pre-training dataset. We refer to the Gemma report for more details \citep{gemma2024}.

We pre-train both RecurrentGemma-2B and RecurrentGemma-9B on 2T tokens. Note that in contrast, Gemma-2B was pre-trained on 3T tokens and Gemma-7B was pre-trained on 6T tokens. Like Gemma, we first train on a large general data mixture, before continuing training on a smaller, higher quality dataset. Like Gemma, we use a subset of the SentencePiece tokenizer \citep{kudo2018sentencepiece}, with a vocabulary size of 256k tokens. Note that, as a consequence of this large vocabulary size, the embedding layer comprises a significant fraction of the total model parameters, as shown in Table \ref{table:hypers}.

\begin{table}[t]
\caption{Relevant formatting control tokens used for both SFT and RLHF of Gemma and RecurrentGemma models.}
    \setlength{\tabcolsep}{6pt}
    \centering
    \footnotesize
    \begin{tabular}{l c c}
    \toprule
    \textbf{Context} & \textbf{Relevant Token} \\
        \midrule
        \scriptsize{User turn} & \texttt{\color{NavyBlue}user} \\
        \midrule
        \scriptsize{Model turn} & \texttt{\color{NavyBlue}model} \\
        \midrule
        \scriptsize{Start of conversation turn} & \texttt{\color{NavyBlue}<start\_of\_turn>} \\
        \midrule
        \scriptsize{End of conversation turn} & \texttt{\color{NavyBlue}<end\_of\_turn>} \\
    \bottomrule
    \end{tabular}
    \label{tab:formatting_tokens}
\end{table}
\begin{table}[t!]
    \caption{Example dialogue with control tokens.}
    \setlength{\tabcolsep}{6pt}
    \centering
    \footnotesize   
    \begin{tabular}{r l}
    \toprule
    \vspace{0.2cm}
    \textbf{User:} & {\color{NavyBlue}\texttt{<start\_of\_turn>user}} \vspace{-0.2cm} \\
    & \texttt{Knock knock.}{\color{NavyBlue}\texttt{<end\_of\_turn>}} \\
    & {\color{NavyBlue}\texttt{<start\_of\_turn>model}} \vspace{0.1cm} \\
    
    \textbf{Model:} & \texttt{Who's there?}{\color{NavyBlue}\texttt{<end\_of\_turn>}} \vspace{0.1cm} \\

    \textbf{User:} & {\color{NavyBlue}\texttt{<start\_of\_turn>user}} \\
    & \texttt{Gemma.}{\color{NavyBlue}\texttt{<end\_of\_turn>}} \\
    & {\color{NavyBlue}\texttt{<start\_of\_turn>model}} \vspace{0.1cm} \\

    \textbf{Model:} & \texttt{Gemma who?}{\color{NavyBlue}\texttt{<end\_of\_turn>}} \vspace{0.1cm} \\

    \bottomrule
    \end{tabular}
    \label{tab:sample_dialogue}
\end{table}
\subsection{Instruction tuning and RLHF}

We follow a similar instruction tuning approach to Gemma \citep{gemma2024}, including a novel RLHF algorithm to fine-tune the model to output responses with high reward. Our instruction tuned model is trained to obey a specific dialogue format, which is defined in Table \ref{tab:formatting_tokens}. For clarity, we give a concrete example in Table \ref{tab:sample_dialogue}.

\section{Evaluation}

We evaluate RecurrentGemma across a broad range of domains, using a combination of automated benchmarks and human evaluation.

\subsection{Automated Benchmarks}

We report the performance of RecurrentGemma on a range of popular downstream evaluations in Table \ref{table:evals}. RecurrentGemma-2B achieves comparable performance to Gemma-2B, even though Gemma-2B was trained on 50$\%$ more tokens. RecurrentGemma-9B achieves comparable performance to Gemma-7B, even though Gemma-7B was trained on $3\times$ more tokens. Note that RecurrentGemma-9B has a similar number of total parameters as Gemma-7B (after accounting for embedding layers).

\subsection{Human Evaluation}

We sent our two final instruction tuned RecurrentGemma models (2B IT and 9B IT) for human evaluation studies against the Mistral 7B v0.2 Instruct model \citep{mistral}. As shown in Table \ref{tab:it}, on a held-out collection of around 1000 prompts oriented toward asking models to follow instructions across creative writing and coding tasks, RecurrentGemma-2B IT achieves a 43.7\% win rate against the larger Mistral 7B model, while RecurrentGemma-9B IT achieves a 59.3\% win rate against the Mistral 7B model.

On a held-out collection of around 400 prompts oriented towards testing basic safety protocols, RecurrentGemma-2B IT achieved a 59.8\% win rate against Mistral 7B v0.2 Instruct model, while RecurrentGemma-9B IT achieved a 59.9\% win rate against Mistral 7B v0.2 Instruct.

\begin{table}[t]
\caption{Win rate of RecurrentGemma-2B IT and RecurrentGemma-9B IT against Mistral 7B v0.2 Instruct, under human evaluation with 95\% confidence intervals. We report a breakdown of wins, ties and losses, and break ties evenly when reporting the final win rate. RecurrentGemma-2B IT 
is surprisingly competitive with the much larger Mistral 7B model, while RecurrentGemma-9B IT performs much better than Mistral 7B v0.2 Instruct on Instruction Following.}
\centering
\small
\begin{tabular}{lrr}
\toprule
Model & Safety & Instruction  \\
 &  &  Following \\
\midrule
\textbf{RecurrentGemma-2B IT} & {59.8\%} &  {43.7\%} \\
\tiny{\textit{95\% Conf. Interval}} & \tiny{[57.1\%, 62.6\%]} & \tiny{[41.8\%, 45.6\%]} \vspace{-0.05cm} \\
\tiny{\textit{Win / Tie / Loss}} & \tiny{47.5\% / 24.6\% / 27.9\%} & \tiny{34.5\% / 18.3\% / 47.2\%} \vspace{0.2cm} \\
%
\textbf{RecurrentGemma-9B IT} & {59.9\%} &  59.3\% \\
\tiny{\textit{95\% Conf. Interval}} & \tiny{[57.1\%, 62.6\%]} & \tiny{[57.4\%, 61.2\%]} \vspace{-0.05cm} \\
\tiny{\textit{Win / Tie / Loss}} & \tiny{44.6\% / 30.7\% / 24.8\%} & \tiny{50.1\% / 18.3\% / 31.5\%} \\
\bottomrule
\end{tabular}
\label{tab:it}
\end{table}

\begin{table*}[t]
\vspace{-2mm}
    \centering
        \caption{Safety academic benchmark results. We provide results for both our pre-trained checkpoint and our instruction tuned variant. For the RealToxicity and Toxigen benchmarks, a lower score is better (indicated by $\downarrow$). For all other benchmarks, a higher score is better (indicated by $\uparrow$).}
    \begin{tabular}{lccccc}
    \toprule
    & & \multicolumn{2}{c}{RecurrentGemma-2B} & \multicolumn{2}{c}{RecurrentGemma-9B} \\
    \cmidrule(l{3pt}r{3pt}){3-4} \cmidrule(l{3pt}r{3pt}){5-6}
    Benchmark & metric & PT & IT & PT & IT \\
      \midrule
    RealToxicity ($\downarrow$) & avg & 9.8 & 7.6 & 10.3 & 8.8 \\  
    BOLD ($\uparrow$) &  & 39.3 & 52.3 & 39.8 & 47.9  \\  
    CrowS-Pairs ($\uparrow$) & top-1 & 41.1 & 43.4 & 38.7 & 39.5 \\  
    BBQ Ambig ($\uparrow$) & top-1 & 62.6 & 71.1 & 95.9 & 67.1 \\  
    BBQ Disambig ($\uparrow$) & top-1 & 58.4 & 50.8 & 78.6 & 78.9 \\  
    Winogender ($\uparrow$) & top-1 & 55.1 & 54.7 & 59.0 & 64.0 \\  
    TruthfulQA ($\uparrow$) & & 35.1 & 42.7 & 38.6 & 47.7 \\  
    Winobias 1\_2 ($\uparrow$) & & 58.4 & 56.4 & 61.5 & 60.6 \\  
    Winobias 2\_2 ($\uparrow$) &  & 90.0 & 75.4 & 90.2 & 90.3 \\  
    Toxigen ($\downarrow$) &  & 56.7 & 50.0 & 58.8 & 64.5 \\  
    \bottomrule
    \end{tabular}
    \label{tab:safety_auto_evals}
\end{table*}

\begin{figure*}[t!]
\centering
\vspace{-1mm}
\begin{subfigure}[t]{0.49\linewidth}
\begin{center}
\includegraphics[height=1.5in]{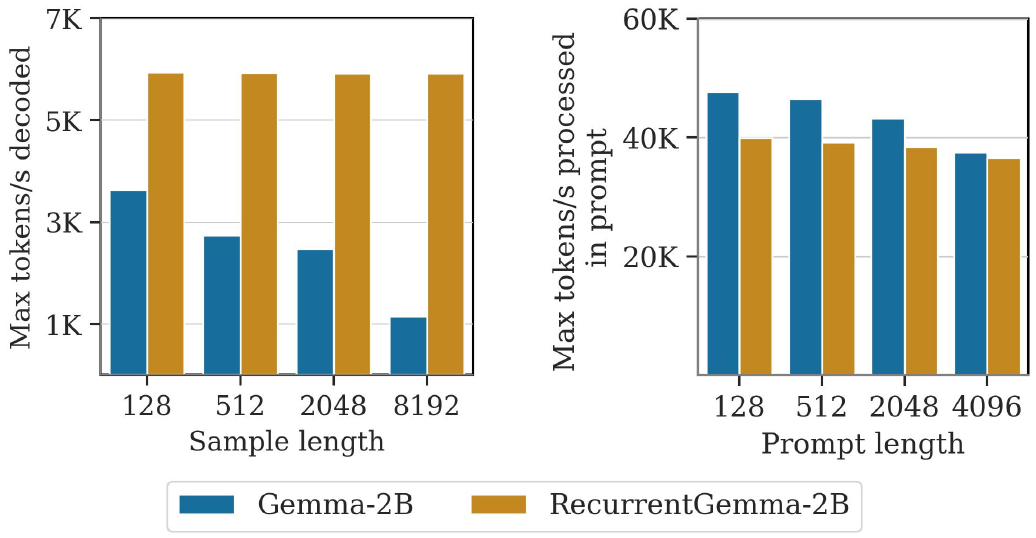}
\caption{Throughput comparison between Gemma-2B and RecurrentGemma-2B on a single TPUv5e.}
\label{fig:inference_2b}
\end{center}
\end{subfigure}
\begin{subfigure}[t]{0.49\linewidth}
\begin{center}
\includegraphics[height=1.5in]{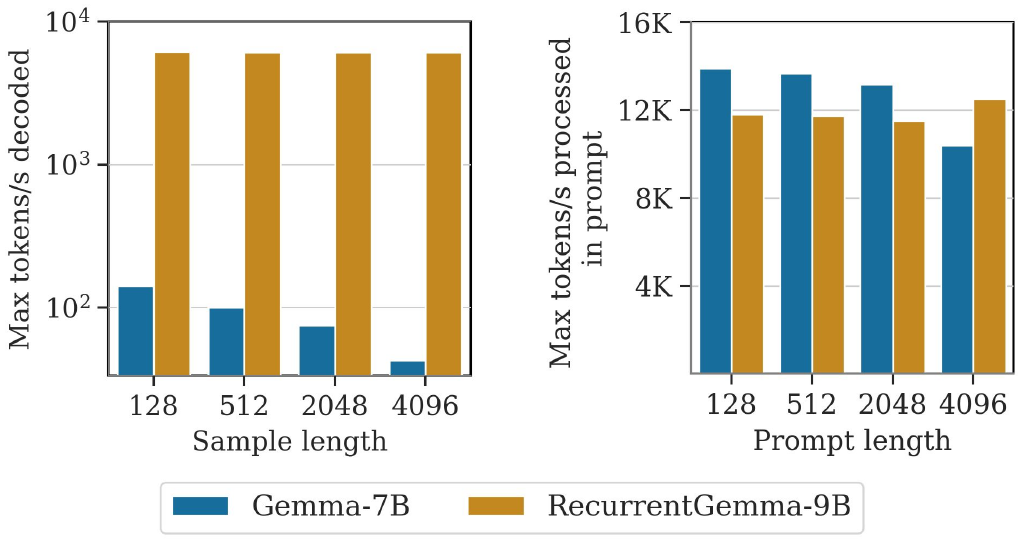}
\caption{Throughput comparison between Gemma-7B and RecurrentGemma-9B on a single TPUv4.}
\label{fig:inference_9b}
\end{center}
\end{subfigure}
\label{fig:inference_figure}
\caption{Maximum tokens per second generated, when sampling sequences of different lengths from a prompt of 2K tokens, and when processing prompts of different lengths to generate the initial state from which to sample, for the RecurrentGemma 2B and 9B models. Both RecurrentGemma models achieve substantially higher sampling throughput than their Gemma counterpart, especially when generating long sequences. A much higher throughput can be achieved when processing input prompts compared to when generating samples, since prompt processing can be efficiently parallelized. RecurrentGemma and Gemma achieve similar prompt processing speeds at both model sizes.}
\end{figure*}

\subsection{Inference Speed Benchmarks}

A key advantage of RecurrentGemma is that it has a significantly smaller state size than transformers on long sequences. Whereas Gemma's KV cache grows proportional to sequence length, RecurrentGemma's state is bounded, and does not increase on sequences longer than the local attention window size of 2K tokens. Inference is typically a memory-bound process for language models \citep{de2024griffin}. Consequently, while the longest sample that can be generated autoregressively by Gemma is limited by the memory available on the host, RecurrentGemma can generate sequences of arbitrary length. Furthermore, the reduced memory requirement also enables RecurrentGemma to perform inference at much larger batch sizes, which amortizes the cost of loading model parameters from host memory into device memory. 


In Figures \ref{fig:inference_2b} and \ref{fig:inference_9b}, we compare the inference throughput achieved by the RecurrentGemma 2B and 9B models to the similarly-sized Gemma models. We first plot the throughput achieved when sampling from a prompt of 2K tokens for a range of generation lengths. The throughput calculates the maximum number of tokens we can sample per second on a single TPUv5e device (in the case of RecurrentGemma-2B) or a single TPUv4 device (in the case of RecurrentGemma-9B). Note that in this plot, we do not account for the time required to process the prompt or the time required to convert the output sequence from a list of token ids into the final text string. RecurrentGemma achieves higher throughput at all sequence lengths considered. The throughput achieved by RecurrentGemma does not reduce as the sequence length increases, while the throughput achieved by Gemma falls as the cache grows. RecurrentGemma-9B achieves particularly large (up to two orders of magnitude) improvements over Gemma-7B as shown in Figure \ref{fig:inference_9b}. We note that this is primarily due to Gemma-7B using Multi-Head Attention, whereas Gemma-2B uses Multi-Query Attention.

For completeness, we also show the throughput achieved when processing input prompts of different lengths. Unlike auto-regressive sampling, the prompt is processed in parallel. Gemma and RecurrentGemma process input prompts at similar speeds. When processing the prompt, both Gemma and RecurrentGemma achieve throughput of roughly 40K tokens per second for the 2B models and roughly 12K tokens per second for the 9B model. By contrast, when sampling, RecurrentGemma achieves throughput of 6K tokens per second, with Gemma substantially slower. Thus, sampling will dominate the total time required, unless the prompt is significantly longer than the desired sample.

Figures \ref{fig:inference_2b} and \ref{fig:inference_9b} were generated using the Flax implementation of RecurrentGemma, which includes a specialized Pallas kernel for execution on TPUs. Users should expect lower throughput when using the Pytorch implementation or when using GPUs. We perform inference for Gemma using a modified version of Gemma's Flax implementation, which we optimized further to improve performance.

\subsection{Responsible Deployment}

We follow the same safety mitigations as described in the Gemma release \citep{gemma2024}. We evaluated our models on standard academic safety benchmarks, as shown in Table 
\ref{tab:safety_auto_evals}, and our final models were also subjected to ethics and safety evaluations by an independent team before release. However, our testing cannot cover all possible use cases of RecurrentGemma, and thus we recommend all users of RecurrentGemma to conduct their own safety testing, specific to their use-case, prior to deployment.

\section{Conclusion}

RecurrentGemma offers the performance of Gemma, while achieving higher throughput during inference, especially on long sequences. We hope that RecurrentGemma will unlock novel applications of highly performant small language models in resource constrained environments.


\clearpage

\input{contributions}


\bibliography{main}

\end{document}

%% file: contributions.tex

\section{Contributions and Acknowledgments}

\noindent\textbf{Griffin Team} \\
Aleksandar Botev\begin{math}\dagger{}\end{math} \\
Soham De\begin{math}\dagger{}\end{math} \\
Samuel L Smith\begin{math}\dagger{}\end{math} \\
Anushan Fernando\begin{math}\dagger{}\end{math} \\
George-Cristian Muraru\begin{math}\dagger{}\end{math} \\
Ruba Haroun\begin{math}\dagger{}\end{math} \\
Leonard Berrada\begin{math}\dagger{}\end{math} \\
Razvan Pascanu
{\let\thefootnote\relax\footnote{\begin{math}\dagger{}\end{math} Joint first authors.}}

\noindent\textbf{RLHF} \\
Pier Giuseppe Sessa \\
Robert Dadashi \\
Léonard Hussenot \\
Johan Ferret \\
Sertan Girgin \\
Olivier Bachem

\noindent\textbf{Gemma Team} \\
Alek Andreev \\
Kathleen Kenealy \\
Thomas Mesnard \\
Cassidy Hardin \\
Surya Bhupatiraju \\
Shreya Pathak \\
Laurent Sifre \\
Morgane Rivière \\
Mihir Sanjay Kale \\
Juliette Love \\
Pouya Tafti \\
Armand Joulin \\
Noah Fiedel \\
Evan Senter

\noindent\textbf{Contributors} \\
Yutian Chen \\
Srivatsan Srinivasan \\
Guillaume Desjardins \\
David Budden \\
Arnaud Doucet \\
Sharad Vikram \\
Adam Paszke \\
Trevor Gale \\
Sebastian Borgeaud \\
Charlie Chen \\
Andy Brock \\
Antonia Paterson \\
Jenny Brennan \\
Meg Risdal \\
Raj Gundluru \\
Nesh Devanathan \\
Paul Mooney \\
Nilay Chauhan \\
Phil Culliton \\
Luiz GUStavo Martins \\
Elisa Bandy \\
David Huntsperger \\
Glenn Cameron \\
Arthur Zucker

\noindent\textbf{Product Management} \\
Tris Warkentin \\
Ludovic Peran

\noindent\textbf{Program Management} \\
Minh Giang

\noindent\textbf{Executive Sponsors} \\
Nando De Frietas \\
Yee Whye Teh \\
Raia Hadsell \\
Zoubin Ghahramani \\
Clément Farabet \\
Koray Kavukcuoglu \\
Demis Hassabis

\noindent\textbf{Acknowledgements} \\
Our work is made possible by the dedication and efforts of numerous teams at Google. We
would like to acknowledge the support from the following teams: Gemini, Gemini Safety, Gemini Infrastructure, Gemini Evaluation, Google Cloud, Google Research Responsible AI and Kaggle.